# Creating Seamless 3D Maps Using Radiance Fields

Sai Tarun Sathyan  ss4005@g.rit.edu
Supervisor: Dr.Thomas Kinsman thomask@rit.edu
Department of Computer Science
Golisano College of Computing and Information Sciences
Rochester Institute of Technology
Rochester, NY 14623

*Abstract*— *It is desirable to create 3D object models and 3D maps from 2D input images for applications such as navigation, virtual tourism, and urban planning. The traditional methods of creating 3D maps, (such as photogrammetry), require a large number of images and odometry. Additionally, traditional methods have difficulty with reflective surfaces and specular reflections; windows and chrome in the scene can be problematic. Google Road View is a familiar application, which uses traditional methods to fuse a collection of 2D input images into the illusion of a 3D map. However, Google Road View does not create an actual 3D object model, only a collection of views. The objective of this work is to create an actual 3D object model using updated techniques. Neural Radiance Fields (NeRF[1]) has emerged as a potential solution, offering the capability to produce more precise and intricate 3D maps. Gaussian Splatting[4] is another contemporary technique. This investigation compares Neural Radiance Fields to Gaussian Splatting, and describes some of their inner workings.  Our primary contribution is a method for improving the results of the 3D reconstructed models. Our results indicate that Gaussian Splatting was superior to the NeRF technique.*

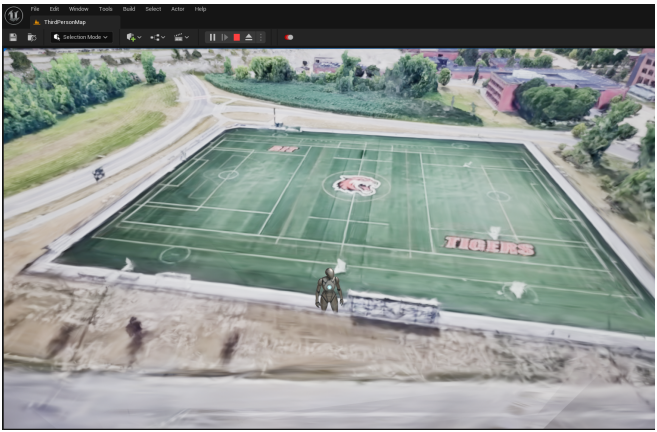

*Fig 0.1: Final Results Preview - RIT Football Field*

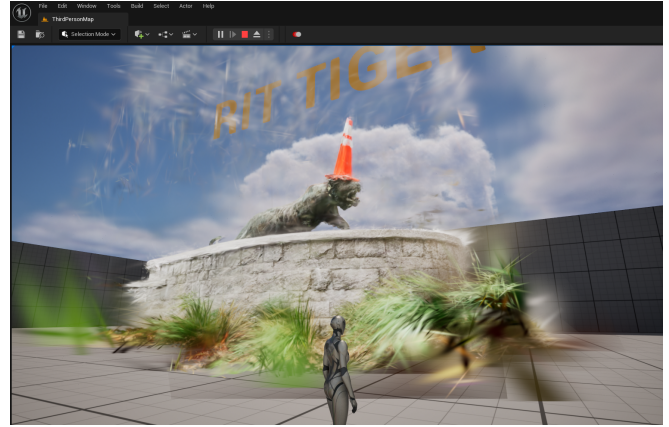

*Fig 0.2: Final Results Preview - RIT tiger statue*

## I. INTRODUCTION

NeRF (Neural Radiance Fields) is a recent development in computer vision that can be used to create 3D maps from a collection of 2D images. NeRF works by learning a mapping from a 5D input (3D position, viewing direction) to a 3D output (color, density). This mapping is represented by a neural network, which is trained on the images [1]. Once the neural network is trained, it can be used to render new views of the scene from any angle.

NeRF has several advantages over traditional methods of creating 3D maps.The project endeavors to investigate the application of Neural Radiance Fields (NeRF) for the generation of cohesive 3D maps representing cities, towns, and college campuses. The methodology involves the systematic collection of a dataset comprising 2D images depicting the target scenes. Subsequently, a NeRF model is trained utilizing this dataset. Once the NeRF model attains proficiency through training, it is employed to render novel perspectives of the scenes from diverse angles. The evaluation of the resultant 3D maps generated by NeRF is conducted, with a subsequent comparative analysis against traditional 3D mapping approaches. NeRF has the potential to revolutionize the way 3D maps are created. NeRF is a powerful tool that can be used to create realistic and seamless 3D maps from a small number of images.  NeRF provides a way to store 3D scenes in a compressed format (as NN weights) with fast query times (I-NGP) and multi-scale photorealistic renders. Gaussian splatting is a rasterization technique for rendering radiance fields in real time. It is different from previous methods in that it represents the radiance field as a set of 3D Gaussians, which makes it more efficient to render and optimize. Gaussian splatting works by first converting the radiance field into a set of 3D Gaussians. Each Gaussian represents a small point in the scene, and the intensity of the Gaussian represents the radiance at that point. The Gaussians are then splatted onto the screen, taking into account their visibility and the angle of the camera [4]. Gaussian splatting is different from previous methods in that it is more efficient to render and optimize. This is because Gaussians are a simple and well-understood mathematical object. They can be rendered quickly using graphics hardware, and they can be optimized efficiently to reduce the number of Gaussians that need to be rendered. Both NerfS and Gaussians have the potential to make 3D maps more accessible and affordable, and to open up new possibilities for applications such as virtual reality and augmented reality. The overarching objective is to establish a robust workflow along with associated scripts to automate the entire process. This initiative aims to facilitate a more accessible and simplified approach for individuals seeking to undertake the creation of seamless 3D maps.

## II. RELATED WORK

### A. "NeRF: Representing Scenes as Neural Radiance Fields" by Mildenhall et al. (2020)

NeRF simplifies 3D reconstruction by modeling scene geometry and appearance within a neural network framework to generate high-quality 3D scenes from 2D images[1]. Despite the original NeRF paper laying the foundation for this novel approach, it did not address real-time constraints, scalability issues, or model optimization potential. This paper conceptualized and developed NeRF, demonstrating its ability to synthesize novel views of a scene from a collection of 2D images, revolutionizing 3D reconstruction. It serves as the fundamental basis upon which subsequent research builds, introducing the concept of neural radiance fields and outlining the mathematical framework for scene representation. Later papers, like "Instant NeRF" and "Zip NeRF," extend this work by addressing practical limitations and efficiency concerns while leveraging the foundational NeRF concept.

### B. "Instant Neural Graphics Primitives with a Multiresolution Hash Encoding" by Thomas Muller et al. (2022)

Instant NGP addresses the real-time limitations of NeRF, enabling practical, interactive applications in augmented reality and robotics [2]. It optimizes NeRF's training and rendering processes, bridging the gap between NeRF's theoretical elegance and practical usability. While it does not fully explore scalability for very large scenes or further model optimization, it is a significant contribution that makes real-time 3D reconstruction achievable.

### C. "Zip-NeRF: Anti-Aliased Grid-Based Neural Radiance Fields" by Jonathan T. Barron et al (2023)

Zip NeRF by Jonathan T. Barron et al. enhances the efficiency and speed of neural radiance fields, making them more scalable and efficient for large-scale scenes[3]. It builds upon NeRF by optimizing the inference process with novel techniques like hierarchical sampling and optimization strategies. Zip-NeRF improves upon the original NeRF paper by getting rid of aberrations in the 3d model and improving the anti-aliasing, depth of field capabilities of NeRF. While Zip NeRF does not delve deeply into addressing dynamic scenes or explore additional use cases beyond real-time 3D reconstruction, it is a significant contribution to the NeRF literature, making large-scale 3D reconstruction more feasible and practical.

### D. 3D Gaussian Splatting for Real-Time Radiance Field Rendering by Bernhard et al (2023)

The paper introduces a new approach to radiance field rendering called 3D Gaussian splatting, which achieves real-time rendering of high-quality radiance fields, even for large and complex scenes [4]. 3D Gaussian splatting is more efficient than traditional radiance field representations because it represents the radiance field as a set of 3D Gaussians, which can be rendered using a simple ray-marching algorithm.This achievement is significant because it makes it possible to render realistic 3D scenes in real time for a wide range of applications, such as augmented reality, virtual reality, and special effects. Overall, the paper is a significant contribution to the field of radiance field rendering. It introduces a new approach that is more efficient and versatile than previous approaches. 3D Gaussian splatting has the potential to revolutionize a wide range of applications.This method significantly advances the field of novel-view synthesis and real-time display of 3D scenes.

## III. SYSTEM FRAMEWORK

In this section we describe the basics for understanding the two primary methods explored for 3D reconstruction, Neural Radiance Fields, and Gaussian Splatting.

*Subsection 1: Neural Radiance Fields and how it works*

Creating a 3D model from 2D images is possible using photogrammetry, neural radiance fields and gaussian splatting. The decision has been made to proceed with the utilization of Neural Radiance Fields and Gaussian Splatting techniques, as they have demonstrated superior efficacy compared to the conventional photogrammetry technique. Photogrammetry, in particular, exhibits limitations in accurately capturing the density of reflective surfaces and transparent objects. When trying to create 3d maps of cities in which most buildings contain reflective surfaces i.e windows it becomes a huge problem to use photogrammetry. The first three papers mentioned in the previous work all depend on neural radiance fields at its core. The final paper relies on gaussian splatting to create 3d models. Neural radiance fields work in a unique way, when it comes to creating 3d models from 2d images. A static scene is represented as a continuous 5D function that outputs the radiance emitted in each direction (θ, φ) at each point (x, y, z) in space, and a density at each point which acts like a differential opacity controlling how much radiance is accumulated by a ray passing through (x, y, z) [1].

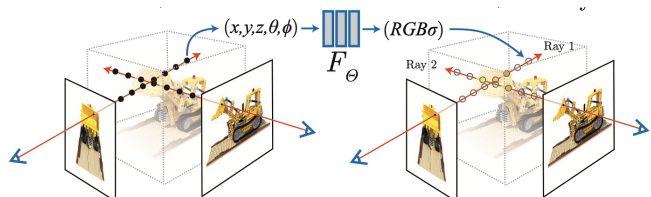

*Fig 1: Neural Radiance Field Input Image Processing*

As seen in Figure 1, NeRF represents a single 5D coordinate (x, y, z, θ, φ) into a single volume density and view-dependent RGB color [1]. NeRF consists of two neural networks. The first one is called the coarse network. In 3d space this shoots out rays from each image and performs hierarchical volume sampling. It samples multiple coarse points spread out through that ray. Each coarse point is evaluated using the coarse network. The coarse network's job is to predict on which particular point on that ray is the volume high and on which point is the volume low. The coarse network rules out which points on this ray are empty. The second neural network used is called the fine network. This neural network focuses on the points at which the volume is greater than zero. It concentrates on the non-empty region and tries to fine tune it. It focuses the computation on non-empty regions and fine tunes what it sees to make the volume density as accurate as possible.

*Subsection 2: Gaussian Splatting and how it works*

The other method that will be explored in this paper is Gaussian splatting. Gaussian splatting allows real-time rendering for scenes captured with multiple photos, and creates the representations with optimization times as fast as the most efficient previous methods for typical real scenes[4].

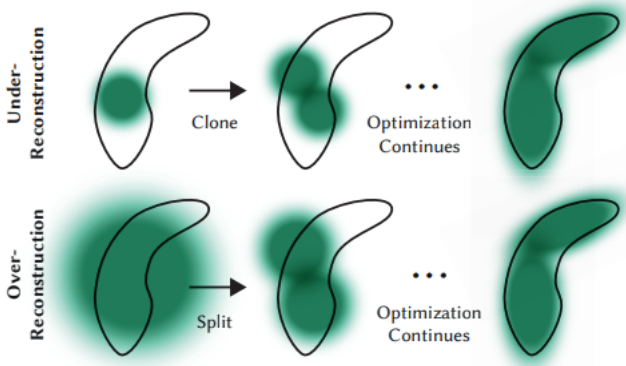

*Fig 2: Gaussian Optimisation*

Gausian splatting starts by using the structure from motion algorithm[SFM][11] to estimate a point cloud from the multiple pictures provided as input. Instead of trying to create a 3D model from the polygons in the point cloud, it converts it into gaussians which take the shape of an ellipsoid. These gaussians now enter an iterative optimization process which fine tunes the gaussian to make it match the original images. In here adaptive density control takes place which either splits up the gaussians if it's too big or joins multiple gaussians if they are too small or transparent. Then the training starts using a process of rasterization, which projects the three dimensional gaussian back onto a two dimensional surface to accurately calculate the depth of each gaussing and the details of each gaussian are compared to the original image and then gaussian gets fine tuned as shown in Fig 2.

Gaussian splatting, unlike photogrammetry or neural radiance fields, does not use ray tracing , path tracing or diffusion. Instead it relies on rasterization techniques for creating an accurate 3D model. This means it converts the underlying data directly into an image. Gaussian splatting requires millions of gaussians to function which requires several gigabytes of virtual ram.

These are the two main techniques which will be employed in order to create seamless 3D maps from 2D images. The "Instant Neural Graphics Primitives with a Multiresolution Hash Encoding" paper improves upon the original NeRF method by making it exponentially faster than the original one. Zip-NeRF is a different approach to NeRF where it projects cones instead of rays through the 2D image, the authors believe this improves depth perception.Because the code for this is unavailable. This project compares instant NGP and Gaussian splatting.

IV. METHODOLOGY

This section delineates the employed methodology for the generation of 3D maps of the scene, utilizing Neural Radiance Fields (NeRF) and 3D Gaussian Splatting. NeRF is a technique that uses a neural network to learn the relationship between 3D positions and the color and density of light at those positions. This allows NeRF to generate realistic views of the scene from any angle. 3D Gaussian splatting is a technique that uses a set of Gaussian functions to represent the 3D volume. It is typically used to render 3D volumes from a fixed viewpoint.

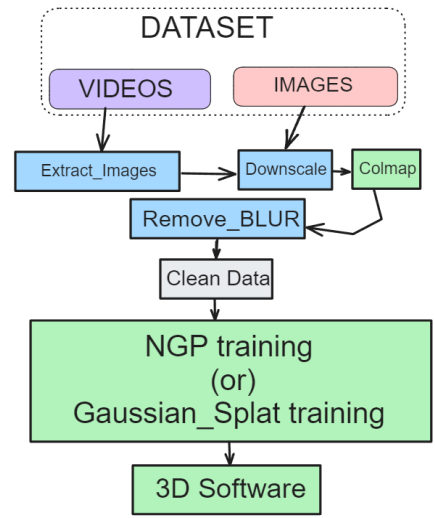

*Fig 3: 3D model creation workflow*

Despite variations in training methods and models, the process of dataset creation remains consistent for both techniques. Drone images of the scene and Google satellite view were employed to capture the requisite data. Subsequently, the workflow transitions to the stages of data cleanup and pre-processing before the prepared data is fed into the training process. Upon completion of the training process for the NeRF neural network and the generation of the 3D Gaussian splatting representation, these models were utilized to render realistic views of the scene from diverse angles. The resulting 3D renders can be exported to third-party 3D model viewing/editing software, enabling manipulation according to specific requirements. A comparative analysis of the rendered views from NeRF and 3D Gaussian splatting was conducted to evaluate the efficacy of the two techniques.

*A. Dataset Creation*

The very first and the most important step of 3d model creation using either NeRF or gaussian splatting is the

Rochester Institute of Technology

dataset capturing and creation technique. These models expect continuous images as raw input. The dataset can be created by capturing video or continuous images and then feeding it to Colmap script which tracks the camera positions for each image and lists it in a json [6].

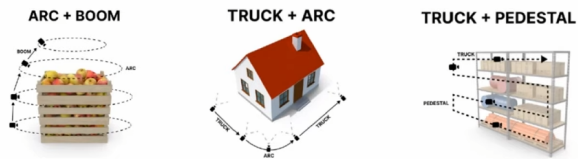

*Fig 4: Data capture techniques for small object big objects and tall object scanning*

As seen in *Fig 4* There are various methods to capture different types of objects, some objects are small enough to be captured with the arc+boom technique which is going in a uniform circle around the object and completing a loop then moving upwards to repeat the same over and over. Whereas some objects are very large and need to be captured using truck+arc which involves going in a straight line and slowly arching around the corner of the object. To address the challenge of capturing exceptionally tall objects, the implementation of a truck-plus-pedestal technique proves advantageous. This method facilitates the comprehensive capture of lateral details, followed by incremental upward movements to replicate the process and ensure complete coverage. All of these methods still require overlapping images to be taken, which is the most important part of dataset creation [5]. Overlapping images help to give Colmap script the context of where the camera is positioned and how it is moving around.

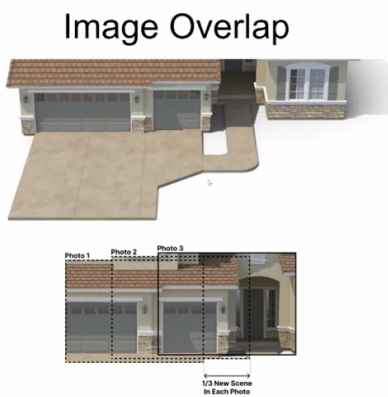

*Fig 5: Image overlap technique*

*Fig 5* shows how images need to be captured in order to Include overlapping photos in the dataset. This ensures that there are no duplications or multiple unknown artifacts in the final 3D model. Another important thing to avoid while capturing data is camera tilt. camera tilt causes chromatic aberrations and blurry 3D models.

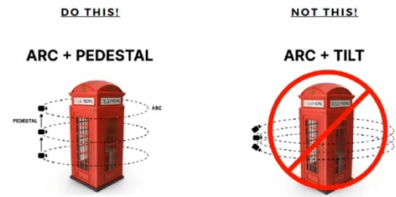

*Fig 6: Avoiding camera tilt*

If in any case you must capture a tall object you can use the arc+boom technique or arc+pedestal technique. Sudden changes in angles during dataset capture results in Colmap not being able to capture the camera movement, hence those images might not make it to the final dataset.

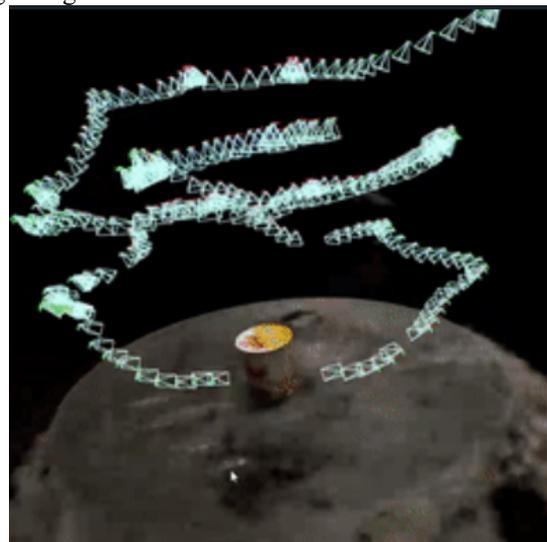

*Fig 7: Avoiding camera tilt*

*Fig.7* shows a good real example of what camera movement should look like while capturing your images for the dataset. Following these techniques will lead to a better 3D model in the 3D reconstruction, any mess up here will cost you a lot of wasted time and resources while training.

B. *Dataset Cleaning*

After your dataset has been created you can directly input this data into training, if you have followed the dataset creation best practices you would get a decent 3d model. But there is still one way to improve this model and it is by removing blurry images out of the dataset. Blurry images obscures details in the objects and confuses NeRF and Gaussian splatting models during training. because blurry images make the model hallucinate that there may be duplicate artifacts in this part of the object or it might lead the model to believe that this part of the object is transparent and the blur is caused by refraction. To avoid these issues blurry images must be removed. But manually reviewing every single image in your dataset and removing blurry images is time consuming and not feasible .To address this challenge, a blur removal script was developed, automating the process. The detection of blurry images was achieved by measuring Laplacian variance as the metric [7].

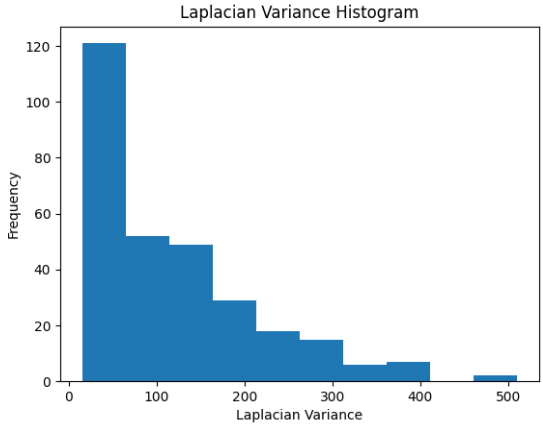

*Fig 8: Dataset - Laplace variance distribution*

Laplacian variance helps to calculate the sharpness of the image and this is done for every image inside our dataset. Next, the mean Laplacian variance is calculated for all images and the standard deviation is calculated as well. The selection of a threshold was based on setting it to any value lower than the mean minus the standard deviation, as determined through the application of the blur removal script. Any image with a Laplacian variance lower than this set threshold is removed from the dataset. Laplacian variance is a measure of the sharpness of an image. It is calculated by computing the Laplacian of the image and then taking the variance of the Laplacian. The Laplacian of an image is a second-order derivative, which means that it highlights high-frequency features in the image.

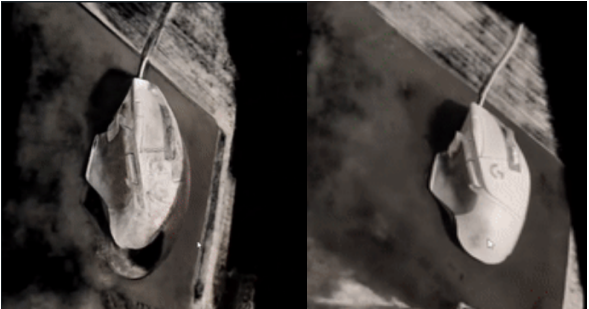

*Fig 9: 3D model: blurry dataset vs non blurry dataset*

As seen in *Fig.9* NeRF can be sensitive to noisy data. If the dataset contains blurry images, NeRF will learn to represent the blurriness in the 3D model. This can result in a 3D model that is not as realistic as it could be. By cleaning the dataset using Laplacian variance, blurry images will be removed from the dataset. This will help NeRF to learn a more accurate representation of the scene, which will result in a more realistic 3D model.

C. *NeRF training*

In NeRF (Neural Radiance Fields), the input to the neural networks comprises a 5D coordinate pair, consisting of 'x,' which represents the 3D spatial location, and 'd,' the 3D Cartesian unit vector denoting the viewing direction. These coordinates are pivotal in predicting two essential components for each input: volume density ($\sigma$), which quantifies the amount of scene information at that location, and directional emitted color (c), representing the color of light emitted from that 3D point in the specified viewing direction.

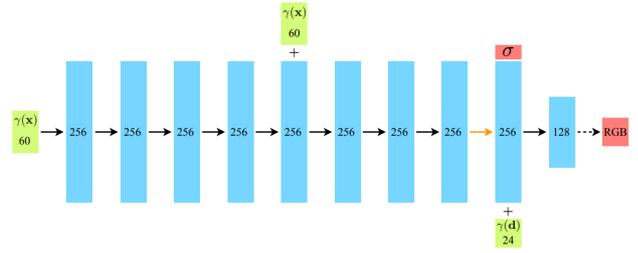

*Fig 10: NeRF network architecture [1]*

The training process involves two neural networks, namely the coarse and fine networks. The coarse network initially processes the 3D coordinates 'x' through a series of fully-connected layers with ReLU activations, employing 256 channels per layer. This operation serves to capture the volume density ($\sigma$) and generates a 256-dimensional feature vector[1].

This feature vector is then combined with the camera ray's viewing direction 'd' and is further processed through an additional fully-connected layer with a ReLU activation and 128 channels, ultimately yielding the view-dependent RGB color (c). The fine network shares a similar architecture but conducts more network queries. It works to refine the representation established by the coarse network, thereby capturing finer details within the scene. In this process, it samples 192 points per ray, utilizing the previously obtained 5D coordinates. This fine-grained approach significantly enhances the quality of the rendered images. NeRF's training and optimization phase involve adjusting the weights of the MLP networks (F$\Theta$) to minimize the disparities between the predicted $\sigma$ and c and the ground truth values extracted from the training images. The network parameters undergo fine-tuning throughout optimization, ensuring that the neural networks effectively model the scene's radiance field[1].

To maintain multiview consistency, NeRF employs an ingenious approach. It confines the prediction of volume density $\sigma$ to be solely a function of the spatial location 'x,' while the prediction of RGB color 'c' depends on both the location and the viewing direction 'd.' This method equips the model to address non-Lambertian effects and effectively capture complex lighting phenomena, including specularities. In summation, NeRF's training process utilizes a pair of neural networks to develop a 5D representation of a scene based on multiple 2D images. These networks collectively consider the spatial structure and appearance of the scene. The coarse network captures the overarching scene structure, and the fine network refines intricate details, culminating in a detailed and view-consistent scene representation[1].

D. *Gaussian Splatting training*

The core of the training process revolves around optimizing the parameters of the 3D Gaussians. This includes refining the position, covariance, $\alpha$, and spherical harmonics (SH) coefficients representing color for each Gaussian. This optimization is performed iteratively by rendering the scene and comparing the result to the training images to adjust Gaussian parameters. The training process needs to handle cases where geometry may be incorrectly placed due to 3D to 2D projection ambiguities, allowing the creation, destruction, or movement of geometry[4].

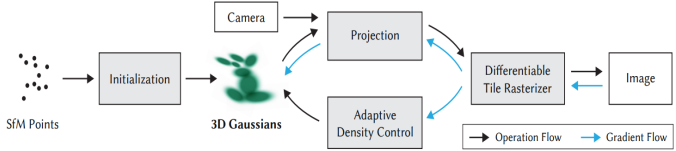

*Fig 11: Gaussian splatting architecture [4] Workflow of the Gaussian splatting method*

The method adaptively controls the number and density of Gaussians over the unit volume to transition from an initial sparse set to a denser set that better represents the scene. This involves densification and removal of Gaussians that are nearly transparent ($\alpha < \epsilon\alpha$) after an initial warm-up phase. Densification focuses on areas with missing geometric features and regions where Gaussians cover large areas in the scene.

Rasterization is a crucial part of the method, where Gaussian splats are rendered efficiently. A tile-based rasterizer is employed to pre-sort primitives for an entire image, which avoids the per-pixel sorting bottleneck. The method utilizes a sorting strategy based on GPU-accelerated radix sort to efficiently order the Gaussians and enable approximate $\alpha$-blending for pixel values. The pixel processing is parallelized, and each pixel accumulates color and $\alpha$ values by traversing the lists of sorted Gaussians. The stopping criterion for rasterization is the saturation of $\alpha$, allowing for an arbitrary number of blended Gaussians to receive gradient updates, which is a notable advantage over previous approaches [4].

The optimization process for Gaussian parameters, along with adaptive control of Gaussians and rasterization, is based on Stochastic Gradient Descent (SGD) techniques, leveraging GPU acceleration for efficient processing. The loss function used combines L1 loss with a D-SSIM term [4]. The entire training process aims to optimize a scene representation that enables high-quality novel view synthesis, starting from sparse SfM points, and employs techniques to handle under-reconstruction and over-reconstruction situations.

It is a multi-step training process that involves adapting a set of 3D Gaussians to represent a scene effectively. It optimizes Gaussian parameters, controls their density adaptively, employs an efficient rasterization technique for rendering, and utilizes stochastic gradient descent for training. This comprehensive approach allows for the creation of a 3D model that can be used for high-quality novel view synthesis.

E.  *Exporting 3D model*

This 3D model created by NeRF or Gaussian splatting could be exported into various 3D softwares. For NeRF the final 3D model created can be exported into blender or Meshlabs. The point cloud can be converted to a mesh to use for various things but the mesh is not perfect, since the point cloud can have various empty spaces filling it up can result in an imperfect and jagged mesh.

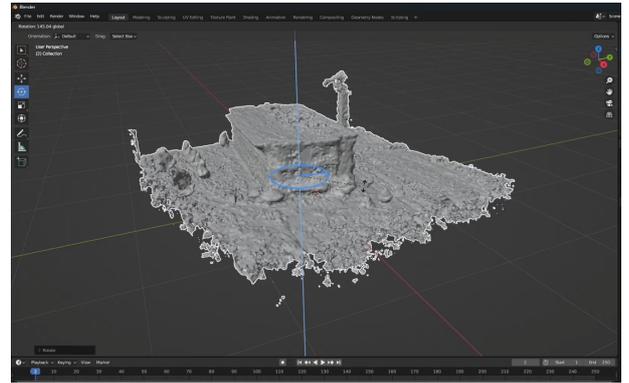

*Fig 12: NeRF model in Blender , displays a crumpled mesh once exported*

The problem with exporting it to blender as seen in *Fig.12* is that it cannot reproduce the colors of the object created by NeRF.

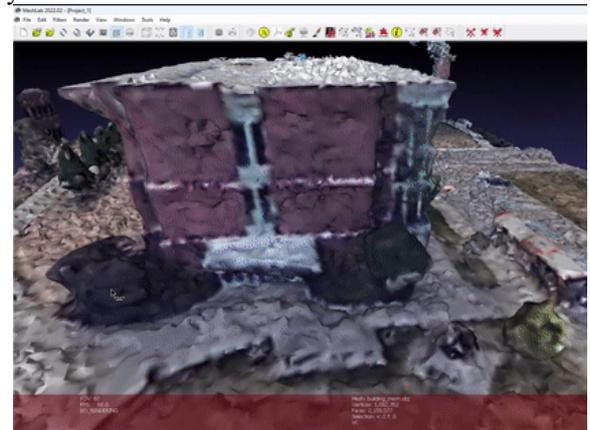

*Fig 13: NeRF model in MeshLabs, similar to blender displays crumpled mesh*

But meshlabs on the other hand is able to reproduce the original colors of the pointcloud. In the case of Gaussian splatting unreal engine[9] plugin [8] can be used to import the 3D model and play around with it.

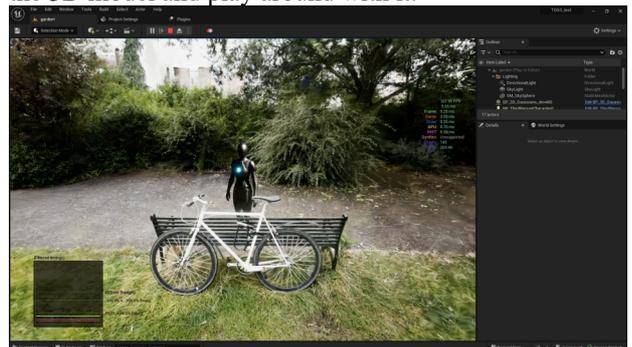

*Fig 14: Gaussian Splatt in Unreal Engine[8][9]*

An advantageous aspect of employing a game engine is the capability to generate an avatar, enabling movement and interaction within the 3D model (as seen in Fig.12) . The mesh object created by 3D gaussian splatting contains color and a cleaner mesh compared to the object produced by NeRF and this might be because gaussian splatting uses ellipsoids in 3D spaces to morph into the object it sees in the image and thus fills up the space in the point cloud, which makes the object have better density distributions compared to the objects produced by NeRF. Gaussian splats can also be exported to unity for free.

## V. EXPERIMENTS

### A. Dataset Capturing Techniques

In accordance with the rules outlined in Section 2.a, an experimentation phase was conducted, involving the capture of diverse objects through the application of various types of camera movements. These are the following data capture techniques that were tested:
1. Keeping the camera parallel to the object and moving around the object while camera is focused on it
2. Moving around the object with the camera facing almost 90° away from it.
3. Moving around the object while the camera has an isometric view of the object

Technique 1 worked in some cases but failed most of the times to capture what was on top of the object or the top down view of it. The effectiveness of the method was observed to be limited to smaller objects. In an attempt to overcome this limitation, Technique 2 was employed, focusing on capturing buildings from a street view. The approach involved navigating around the building while ensuring it remained within the peripheral view of the camera. The end result was the 3D model being unable to map out the building properly in 3D space instead it created a 3D model of the building in a straight line rather than 4 sides of the building.Enhanced testing opportunities can be explored by employing 360° equipment, potentially yielding improved results in this context. However, it is worth noting that such equipment was not available during the conducted experiments. Technique 3 proved to be the most useful method of data capture since it produced the most detailed 3D models as the end result. 3D models created using technique 3 as the data capture technique had the best 3D models in terms of details and accuracy.

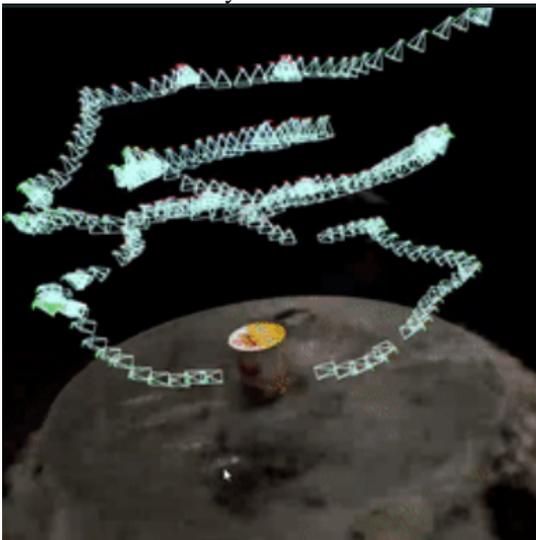

Fig 15: Camera movement for technique 3. This shows the camera motion for Arc+Boom technique, where the camera is mounted on a boom

As seen in *Fig.15* The important thing to keep in mind is to never change the camera angle. A 45° angle was maintained throughout the capture and moved in a circular motion around the object. Going around the object in a smooth motion, starting close to it and moving further away and higher up in the 2nd and 3rd rotation helps capture all details necessary for the 3d model recreation.

### B. Comparing NeRF and Gaussian Splatting

NeRFs and Gaussian splatting both have different 3D model recreation methods. In short, NeRFs produce a point cloud with colored points and Gaussian splatting produces 3D ellipsoid fields with color and transparency. Upon examination within their respective graphical user interfaces (GUI), both 3D models appear nearly identical. However, when these models are exported to third-party 3D modeling software, the genuine disparities between the final results become apparent.

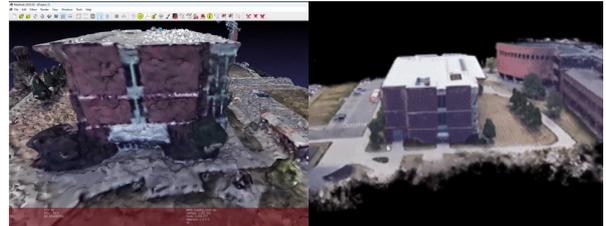

Fig 16: NeRF export compared with GUI preview

Here in *Fig.16* it can be seen that the 3D model produced by NeRF looks fine in the NeRF GUI but when exported to meshlabs it loses a lot of detail and the model looks extremely choppy. This is because the point cloud itself is not perfect even i.e the point cloud density is not a perfect representation of what is seen in real life. Due to the parallax effect it is observed as a perfect 3D model in the NeRF GUI. During the exportation process to software such as MeshLab or Blender, a thin mesh layer is applied over the object to ensure uniformity in the 3D mesh. It is at this stage that the genuine flaws in the 3D model produced by NeRF become apparent.

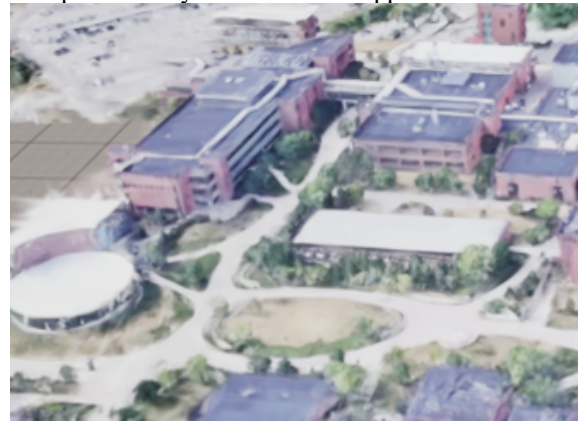

Fig 17: Gaussian Splatting export

Upon exporting the Gaussian splatting model to the Unreal Engine, the resulting representation is observed as depicted in Fig.17.

### C. 3D model comparison and scaling

Following the assessment of individual object scans using both Gaussian splatting and NeRF, the subsequent phase involved investigating the capacity of these techniques to accurately capture multiple objects simultaneously. To find out the upper limit of these scans, google satellite view[10] was used to capture multiple buildings in RIT campus and feed the training data to both the NeRF and Gaussian splatting model.These were the results of both the models outputs starting from smaller single object scans to much wider multi object scans (same dataset was used to train both):

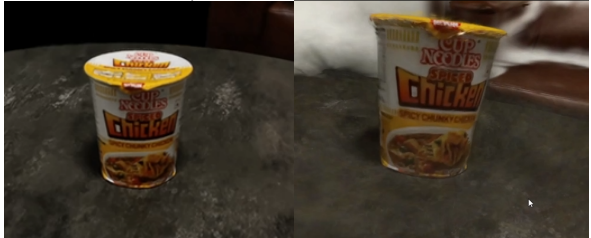

*Fig 18: Ramenbox NeRF vs Gaussian Splatting*

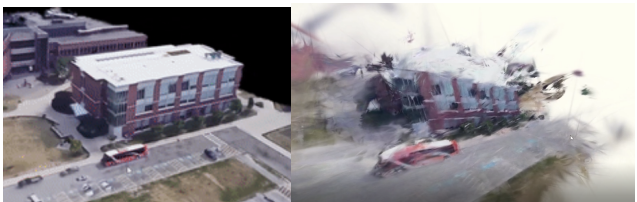

*Fig 19: One Building NeRF vs Gaussian Splatting*

The problem seen in Fig.19 is that the building is not being accurately reproduced in the gaussian splatting model. Upon scrutinizing the training data to identify the source of the issue, it was discerned that the dataset contained slight variations in camera tilt throughout the capture process. This, in turn, led to significant distortions in the 3D model during the training of Gaussian splats. This was an interesting insight to take note: Gaussian splatting is much more sensitive to camera tilts and this might cause the end result to be extremely distorted. Upon the removal of images from the dataset that exhibited camera tilt, the subsequent output for the Gaussian splatting training process was observed:

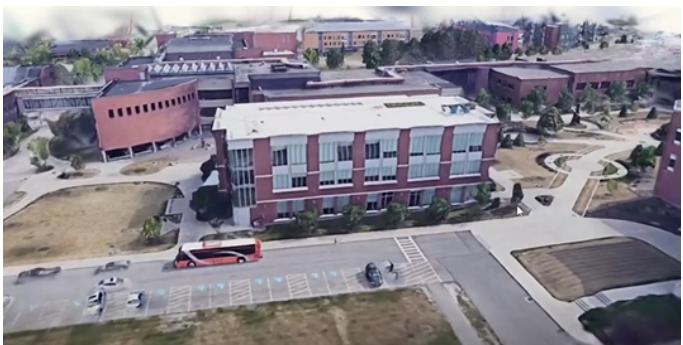

*Fig 20:One Building Gaussian Splatting model clean dataset*

Next for the multi object scan these were the results:

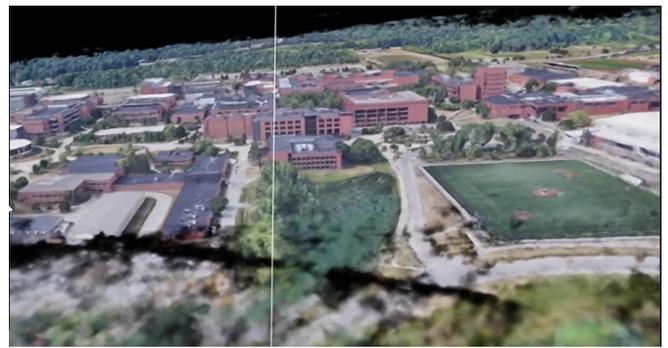

*Fig 21.1 :Multi Building **NeRF Map***

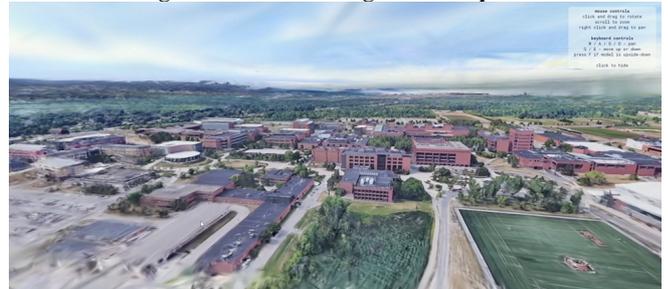

*Fig 21.2:Multi **Gaussian Splatting Map***

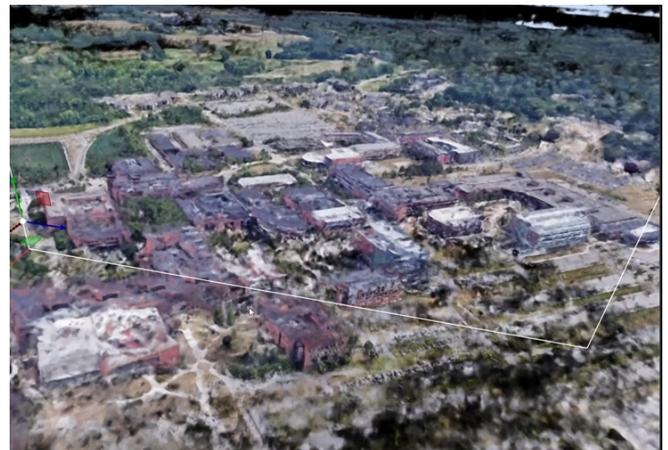

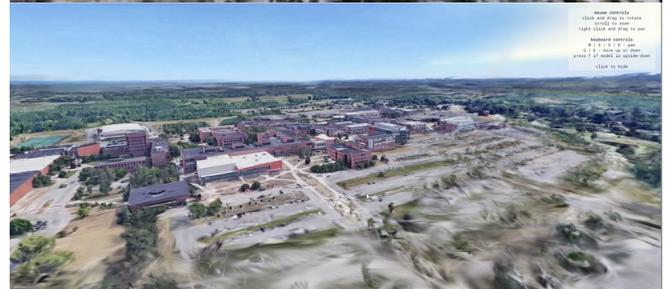

*Fig 22:Multi Building NeRF vs Gaussian Splatting*

In the multi-object scan test, Gaussian splatting has outperformed NeRFs render. Upon examination of the buildings from an isometric view, as depicted in Fig.22, it is evident that NeRF encountered challenges in accurately replicating the density of the point cloud at several locations. This limitation led to the presence of void spaces in various areas. The Gaussian splatting model seems to have handled these areas a bit better, even though these areas contain rough and wide Gaussian patches, it is a better result than having an empty hole in the ground.

Rochester Institute of Technology
8

| Comparison (for one building scan) | Memory efficiency | Resource consumption | Accuracy | Input data | Training speed |
|---|---|---|---|---|---|
| NeRF | 176mb | 8gb vram | 90% | standard continuous images | 1 min |
| Gaussian Splatting | 256mb | 24gb vram | 95% | standard continuous images, but sensitive to camera shakes | 5 min |

*Table 1: Gaussian Splatting vs NeRF (one building scan)*

## VI. FUTURE WORK

For prospective project expansions, considerations involve the development of an Unreal Engine script aimed at seamlessly incorporating Gaussian splats as grids and positioning them upon import. Such a script could significantly streamline the workflow associated with the creation of large-scale maps. Additionally, there is an interest in exploring stable diffusion generative AI techniques to enhance the visual aesthetics of the final 3D models produced.

## VII. RESULTS AND DISCUSSION

Following a thorough evaluation of the performance of both NeRF and Gaussian splatting models, the decision was made to proceed with the Gaussian splatting model for the purpose of generating 3D maps. Here are some pics of the end results of the 3D map generated from training of 1000+ images using gaussian splatting technique, exported to unreal engine:

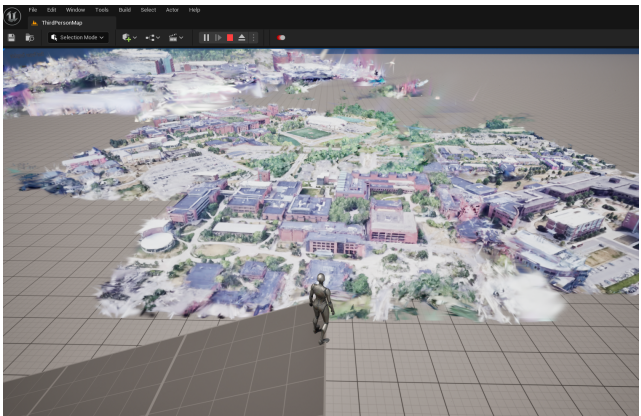

*Fig 23.1: Top down view of RIT Map in Unreal Engine*

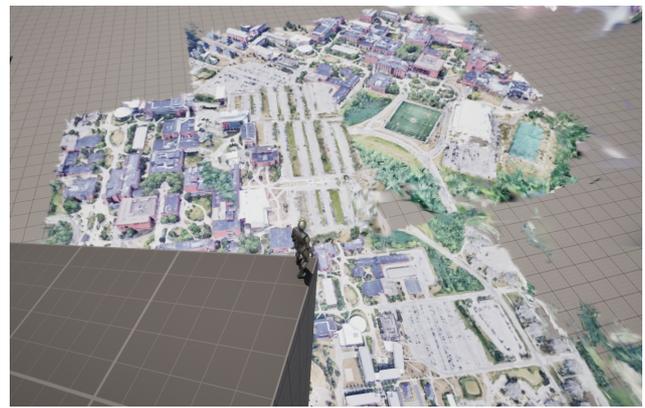

*Fig 23.2: Top down view of RIT Map in Unreal Engine*

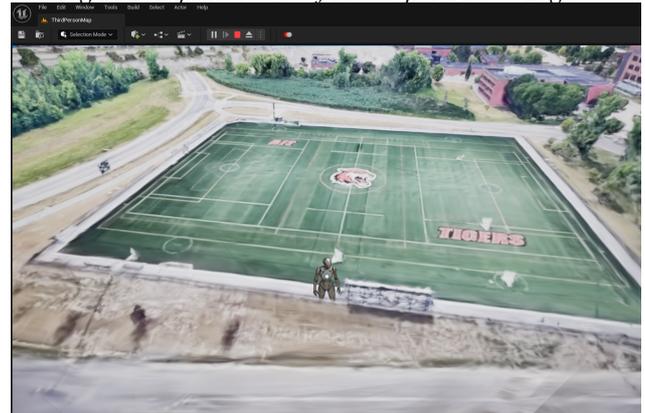

*Fig 23.3: Running around RIT Map in Unreal Engine*

These assets in Fig.23 were created using google satellite view as input data and trained on the gaussian splatting model. Owing to the imposed drone flying restrictions on the university campus, recording the requisite videos essential for training purposes was unattainable.

The exports from NeRF render looked subpar to the point cloud that was displayed on the preview window. This might be due to the NeRF model producing a very sparse point cloud which looks bad when a final mesh is applied on top of it, here is a real world object comparison with NeRF export and Gaussian splatting export:

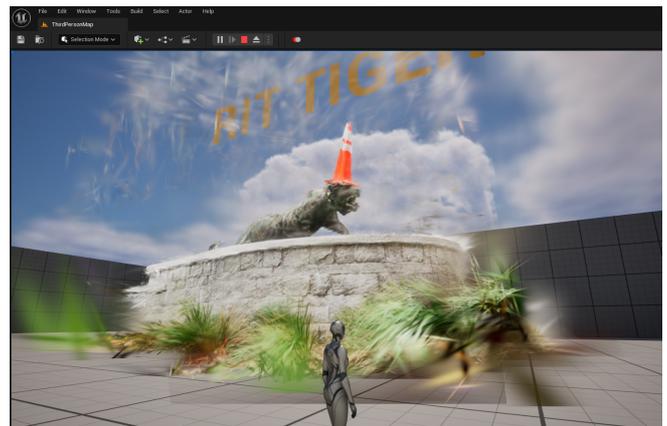

*Fig 24:RIT tiger - Gaussian Splatting model*

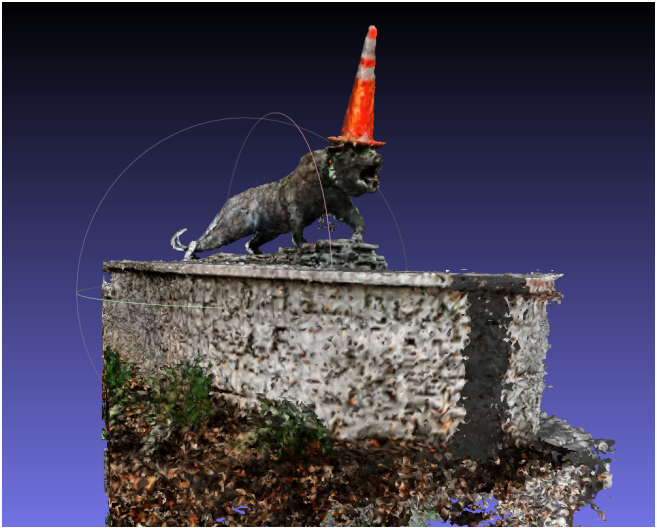

*Fig 25:RIT tiger - NeRF model*

## VIII. CONCLUSION

The evaluation compared NeRF and Gaussian Splatting for 3D object reconstruction from video sets. A key contribution was the use of a pre-processing technique to select high-quality input frames, leading to sharper outputs for both methods. Notably, Gaussian Splatting outperformed NeRF in achieving the sharpest 3D reconstructions. This emphasizes the effectiveness of Gaussian Splatting for high-fidelity 3D reconstruction, particularly when combined with strategic data pre-processing. In the end Gaussian Splatting technique proved to me more accurate but also used more computing resources than NeRF.

## IX. ACKNOWLEDGEMENTS

I would like to express my sincere gratitude to Professor Thomas Kinsman for his invaluable guidance and support throughout this project. His expertise, patience, and enthusiasm were instrumental in shaping my understanding of the subject matter. His insights and feedback have been invaluable in refining my research and developing my critical thinking skills.